\documentclass{article}
\usepackage{spconf,amsmath,graphicx}
\usepackage{amsfonts,amssymb}
\usepackage{xcolor}
\usepackage{multirow}
\usepackage[absolute,overlay]{textpos}

\title{Query-by-Example Keyword Spotting system using Multi-head Attention and Softtriple Loss}
\name{Jinmiao Huang$^{1^*}$ \thanks{*Authors contributed equally} \qquad Waseem Gharbieh$^{1^*}$ \qquad Han Suk Shim$^{2}$ \qquad Eugene Kim$^{2}$}
\address{$^{1}$ Toronto AI Lab, LG Electronics, Toronto, Canada\\
    $^{2}$ Artificial Intelligence Lab, LG Electronics, Seoul, Korea}

\begin{document}

\maketitle

% \begin{textblock*}{17.8cm}(20mm,26.5cm) % {block width} (coords) 
\begin{textblock*}{17.8cm}(20mm,0.2cm) % {block width} (coords) 
  \footnotesize$\copyright$ 2021 IEEE. Personal use of this material is permitted. Permission from IEEE must be obtained for all other uses, in any current or future media, including reprinting/republishing this material for advertising or promotional purposes, creating new collective works, for resale or redistribution to servers or lists, or reuse of any copyrighted component of this work in other works.
\end{textblock*}

\begin{abstract}

This paper proposes a neural network architecture for tackling the query-by-example user-defined keyword spotting task. A multi-head attention module is added on top of a multi-layered GRU for effective feature extraction, and a normalized multi-head attention module is proposed for feature aggregation. We also adopt the softtriple loss - a combination of triplet loss and softmax loss - and showcase its effectiveness. We demonstrate the performance of our model on internal datasets with different languages and the public Hey-Snips dataset. We compare the performance of our model to a baseline system and conduct an ablation study to show the benefit of each component in our architecture. The proposed work shows solid performance while preserving simplicity.

\end{abstract}
\section{Introduction}
\label{sec:intro}
A keyword spotting (KWS) system is a significant starting point of a conversation when users talk to voice assistants. Most devices adopting voice assistants use a pre-defined keyword such as ``Hey Siri", ``Alexa" or ``OK Google". Although user-defined wake-up-words (keywords) can be attractive as it is one way to personalize devices, its application is challenging because custom utterances can be out of the training distribution. Furthermore, low latency and small memory footprints need to be addressed in order to effectively run user-defined KWS on devices. 

Although recent neural network approaches have improved the performance on the pre-defined keyword spotting task \cite{chen2014small, tang2018deep}, their application to the user-defined keyword task has remained elusive. To improve the performance on the user-defined keyword task, some researchers have used the Large Vocabulary Continuous Speech Recognition system \cite{miller2007rapid}. These systems usually show performance degradation when the keywords are out of vocabulary and require a lot of computational overhead.

Recently, using query-by-example (QbyE) has been recognized as a more promising approach for tackling user-defined KWS. In a QbyE system, the user enrolls the desired wake word by recording a few examples, the KWS system then compares the incoming audio with these examples to detect the wake word. Early QbyE approaches used dynamic time warping (DTW) to determine the similarity between keyword samples and test utterances \cite{hazen2009query, zhang2009unsupervised}. Recent QbyE work has used neural networks to map variable-duration audio signals to fixed-length vectors, which showed great speed and performance advantages over DTW-based systems. \cite{chen2015query} trained an LSTM model for token classification, and took the last few time steps of the hidden layer as the embedding. In the spoken term detection task, \cite{chung2016audio} trained a denoising seq2seq autoencoder in an unsupervised manner to obtain a fixed-length representation. Similarly, \cite{audhkhasi2017end} used a seq2seq GRU autoencoder to encode the audio, and a CNN-RNN language model to encode the characters. In addition, \cite{lugosch2018donut} used Connectionist Temporal Classification beam search to produce a hypothesis set based on the posteriors of ASR. \cite{he2017streaming} used an RNN transducer model for predicting subword units. \cite{settle2016discriminative} uses a Siamese network with triplet hinge loss. A number of attention-based models have also been proposed to tackle the QbyE task \cite{ao2018query, rahman2018attention}.

In this paper, we propose a neural architecture that tackles the user-defined KWS task. Our contributions are: (1) We use multi-head attention \cite{vaswani2017attention} together with RNNs to effectively extract the features while maintaining a small model size. (2) A simple normalized multi-head attention layer is proposed to aggregate the information from the feature extractor. (3) We used the softtriple loss \cite{qian2019softtriple}, which combines the advantages of softmax loss and triplet loss to ensure inter-class separation in the embedding space. (4) We demonstrate the performance of our model on the Hey-Snips \cite{coucke2019efficient} public dataset, which can be used as a baseline for future research.

\section{Keyword Spotting System}
\label{sec:system}
% The ``encoder-decoder" structure can be naturally applied to tackle the QbyE problem. In our system, the  encoder's role is to map input signals to an embedding vector, and the decoder's role is to maximize the inter-class (different classes) distances, and minimize the intra-class (same class) distances given the input embedding. At inference time, the distances between the embeddings of the enrollments and the queries are computed. Therefore, the decoder network is completely dropped. The system architecture is shown in Fig.\ref{fig:encoder-decoder} 

\begin{figure}[htb]
  \centering
  \includegraphics[width=8.5cm]{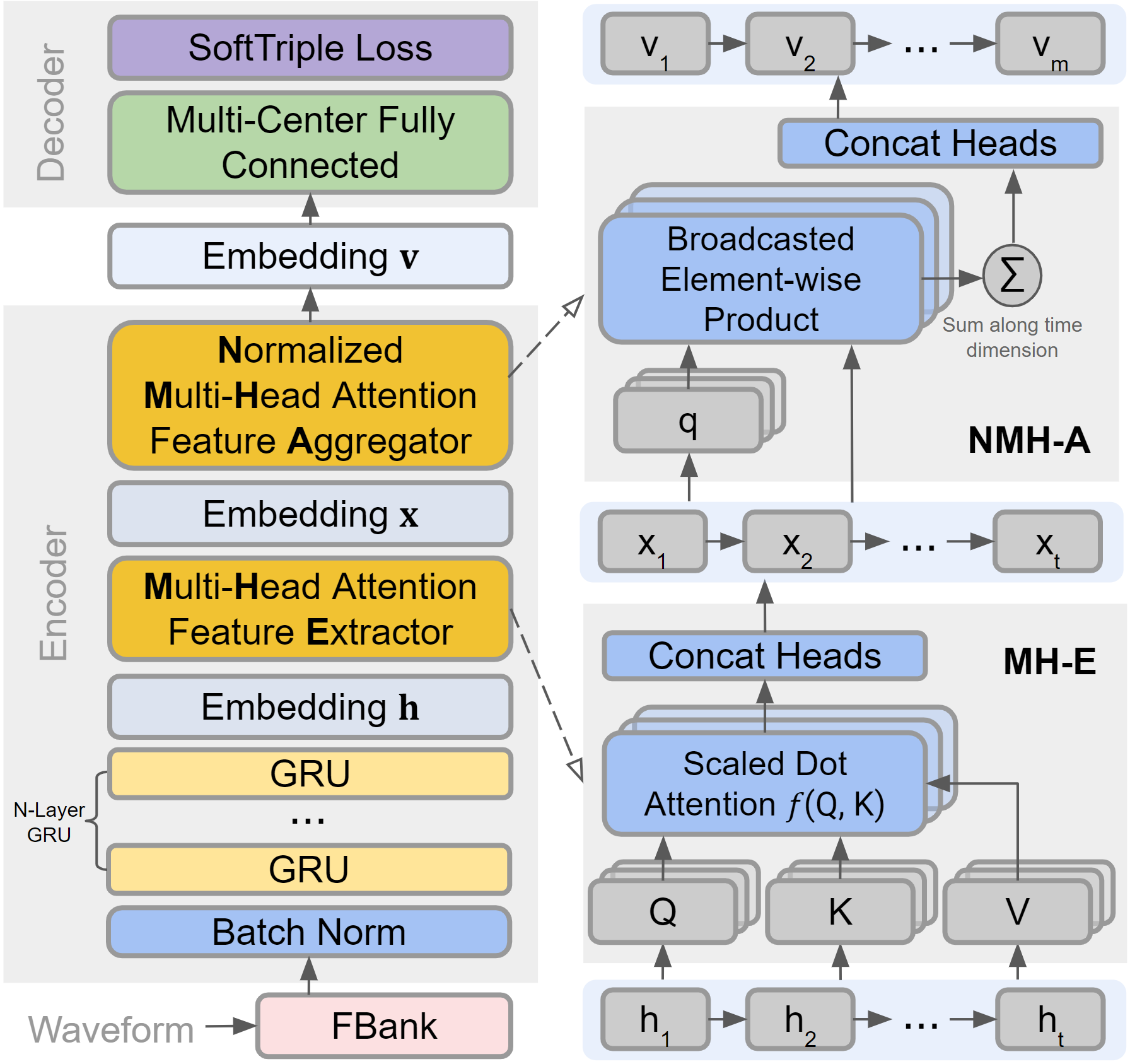}
%  \vspace{2.0cm}
\caption{Multi-Head Attention Encoder-Decoder Architecture}
\label{fig:encoder-decoder}
\end{figure}

We use the ``encoder-decoder" structure to tackle the QbyE problem. Fig.\ref{fig:encoder-decoder} shows the system architecture. The  encoder's role is to map input signals to an embedding vector, and the decoder's role is to maximize the inter-class (different classes) distances and minimize the intra-class (same class) distances given the input embeddings. At inference time, the distances between the embeddings of the enrollments and the queries are computed, and the decoder is completely dropped.

\subsection{Base Encoder}
For feature extraction, We compute 160-dimensional Mel filterbank (FBank) features of the input waveform using a 25 ms window and a stride of 12 ms. In the base encoder, a multi-layer GRU is used to extract the embeddings from the FBank features after passing through a batch normalization layer. The output of the GRUs is a sequence of vectors $\mathbf{h} = [h_1, h_2, \ldots, h_t]$, each vector $h_i\in\mathbb{R}^{n}$ represents the hidden state at time $i$ from the last GRU layer. In this section, we use $n$ as the number of features, $t$ as the number of time steps, and $m$ as the number of heads in our attention mechanism.  

\subsection{Multi-Head Attention Mechanism}
An attention mechanism typically computes an alignment score between a query vector and a source sequence. In our system, we first apply the multi-head attention mechanism \cite{vaswani2017attention} on top of the base encoder which serves as an additional feature extractor, we then create a simple normalized multi-head attention mechanism to serve as a feature aggregator to reduce the dimensionality of the feature space.

\paragraph*{Multi-Head Feature Extractor (MH-E):}
We use Multi-head self-attention from  \cite{vaswani2017attention} to project the input into multiple subspaces to capture the relationship between the features at different positions. We define the multi-head mechanism as $\left\{\mathbf{Q}_{j} = \mathbf{W}_{j}^{q}\mathbf{h}, \mathbf{K}_{j}=\mathbf{W}_{j}^{k}\mathbf{h}, \mathbf{V}_{j}=\mathbf{W}_{j}^{v}\mathbf{h}\right\} \in \mathbb{R}^{t \times d}$, where $\mathbf{Q}_{j}$, $\mathbf{K}_{j}$, $\mathbf{V}_{j}$ are the query, key, and value representations of the input embedding $\mathbf{h}$ on the $j$th head, and $\mathbf{W}_{j}^{q}, \mathbf{W}_{j}^{k}, \mathbf{W}_{j}^{v} \in \mathbb{R}^{n \times d}$ are the weight matrices associated with the $j$th head, and $d$ is the dimension of the projected subspace. We use the scaled dot attention between $\mathbf{Q}_j$ and $\mathbf{K}_j$ to generate a score $\mathbf{a}_j \in \mathbb{R}^{t\times t}$. We then multiply $\mathbf{a}_j$ by $\mathbf{V}_j$ to get $\mathbf{x}_j \in \mathbb{R}^{t\times d}$. The final output $\mathbf{x} \in \mathbb{R}^{t\times md}$ is the concatenation of  $\mathbf{x}_j$ from each attention head.

% \begin{equation}
% \label{eq:multihead}
% \mathbf{x}_{j} = \mathbf{a}_j\mathbf{V}_j, \quad  \mathbf{a}_j=\operatorname{softmax} \left(\frac{\mathbf{Q}_j \mathbf{K}_{j}^{\top}}{\sqrt{d}}\right)
% \end{equation}

% To control the model parameter size, we fixed $d$ to $n/m$, and tuned the hyper parameter $m$. Our experiments showed that with fixed $m\times d$, models generally perform better when $m$ is large.

\paragraph*{Normalized Multi-Head Feature Aggregator (NMH-A):}
Since $\mathbf{x}$ is usually large, some previous work has sought to take the last time step \cite{chung2016audio} or the last k time steps \cite{chen2015query} to reduce the size of the embedding vector. \cite{ao2018query} and \cite{rahman2018attention} make use of attention to get a weighted sum of the feature vector. They use the hidden vector $h_t$ from the last time step as the query, and the $\mathbf{x}$ from all the hidden states as the source sequence to compute an alignment score $\mathbf{\alpha}$, to obtain the final embedding.

In our system, we develop a simple normalized multi-head attention mechanism. We first define a weight matrix $\mathbf{W}\in\mathbb{R}^{n \times m}$. The query vector for each head is defined as $\boldsymbol{q}_j =\mathbf{W}_j^{\top}\mathbf{x}$. The intuition is that: $\mathbf{W}_j^{\top}\mathbf{x} =\|\mathbf{W}_j\|\|\mathbf{x}\| \cos \left(\theta_j\right)$ where $\theta_j$ is the angle between embedding vector $x$ and the weight on the $j$th attention head. This indicates that both, the norm and the angle, contribute to the projection. To develop effective feature learning for each head, we use $L_2$ normalization on $\mathbf{W}$ to produce $\mathbf{W^*}$ such that $\|\mathbf{W^*}\|_2 = 1$. This ensures that the attention score will not be dominated by the norm of the weight matrix $\mathbf{W}$. 

In the next step, we apply a softmax function to normalize $\boldsymbol{q}$ along the time dimension to create a multi-head alignment vector (Eq.\ref{eq:attention}). We then multiply $\boldsymbol{q}$ and $\mathbf{x}$ to create a weighting vector, and sum the output along the time dimension to create a multi-head weighted sum vector $\boldsymbol{v} = [v_1, v_2, \dots, v_m]$ where $v_i$ represents the weighted sum at each head. The final output vector is the concatenation of $v_i$ along the feature dimension. Specifically, the attention model is implemented as follows:

\begin{equation}
\label{eq:attention}
q_{ij} = \frac{\exp \left( \mathbf{W}_j^* \mathbf{x}_i \right) } {\sum_{k=1}^t \exp \left( \mathbf{W}_k^* \mathbf{x}_k\right)}
\end{equation}
\begin{equation}
\boldsymbol{v}_j = \sum_{i=1}^t q_{ij}\mathbf{x}_i
\end{equation}
% \begin{equation}
% \label{eq:attention_output}
% \boldsymbol{u} = [v_1;v_2;\dots;v_m] \\
% \end{equation}

where $\mathbf{x}\in\mathbb{R}^{t \times n}$, $\mathbf{W}^*_j\in\mathbb{R}^{n}$, $\boldsymbol{q} \in \mathbb{R}^{t \times m }$, $\boldsymbol{v} \in \mathbb{R}^{m \times n}$. $\mathbf{W}^*$ is a learnable parameter.

% Compared to \cite{vaswani2017attention}, our attention uses one weight matrix, and project the original input matrix into a smaller space, which serves as feature aggregation rather than feature extraction used in \cite{vaswani2017attention} 

% \cite{tang2018deep} applied residual network with dilated convolutions to explore CNN-based model for small-footprint keyword spotting. \textcolor{green}{Parameter size comparison with them}

\subsection{Decoder with Softtriple Loss}
The QbyE system can be cast as an optimization problem with triplet constraints, where the objective is to minimize the distance between the embedding of a specific enrolled keyword and the embeddings of the same keyword while maximizing the distance with respect to other keywords. 

% We first implemented triplet loss follow Eq.\ref{eq:triplet} from \cite{schroff2015facenet}. To accelerate the training, we add the batch hard miner \cite{hermans2017defense} and a customized miner based on mini-batch Levenstein distance. However, for the same encoder network structure, our triplet algorithm cannot match the performance of the softmax classification loss.
% \begin{equation}
%     \label{eq:triplet}
% \forall \mathbf{x}_i^a, \mathbf{x}_i^p, \mathbf{x}_i^n, \quad\left\|\mathbf{x}_i^a-\mathbf{x}_i^p\right\|_{2}^{2}-\left\|\mathbf{x}_i^a-\mathbf{x}_i^n\right\|_{2}^{2} \geq \delta
% \end{equation}

% \cite{liu2016large} purpose using cosine distance as the classification score in Softmax by the fact that the last fully connected layer $f_{y_{i}}=\boldsymbol{W}_{y_{i}}^{T} \boldsymbol{x}_{i}$ (omit the bias $b$ for simplicity) can be also formulated as $f_{j}=\left\|\boldsymbol{W}_{j}\right\|\left\|\boldsymbol{x}_{i}\right\| \cos \left(\theta_{j}\right)$ where
% $\theta_{j}\left(0 \leq \theta_{j} \leq \pi\right)$ is the angle between the vector $\boldsymbol{W}_{j}$ and $\boldsymbol{x}_{i} .$ Thus the softmax loss becomes

% \begin{equation}
% \label{eq:l-softmax}
% L_{i}=-\log \left(\frac{e^{\left\|\boldsymbol{W}_{y_{i}}\right\|\left\|\boldsymbol{x}_{i}\right\| \cos \left(\theta_{y_{i}}\right)}}{\sum_{j} e^{\left\|\boldsymbol{W}_{j}\right\|\left\|\boldsymbol{x}_{i}\right\| \cos \left(\theta_{j}\right)}}\right)
% \end{equation}

One way to achieve this objective is by using triplet loss. However, an effective sampling strategy over the mini-batch triplets is essential to learn the embeddings efficiently \cite{hermans2017defense}. Another way is by combining triplet loss with softmax loss to produce softtriple loss \cite{qian2019softtriple} where each class is associated with multiple centers, and a relaxed version of similarity between the example $\mathbf{x}_{i}$ and the class $c$ can be defined as in Eq.\ref{eq:similarity}. It can also be proven that optimizing the triplets consisting of centers with a margin $\delta$ can preserve the large margin property on the original triplet constraints as in Eq.\ref{eq:softtriple}. Please refer to the original paper for details.

\begin{equation}
\label{eq:similarity}
\mathcal{S}_{i, c}^{\prime}=\sum_{k} \frac{\exp \left(\frac{1}{\gamma} \mathbf{x}_{i}^{\top} \mathbf{w}_{c}^{k}\right)}{\sum_{k} \exp \left(\frac{1}{\gamma} \mathbf{x}_{i}^{\top} \mathbf{w}_{c}^{k}\right)} \mathbf{x}_{i}^{\top} \mathbf{w}_{c}^{k}
\end{equation}

\begin{equation}
\label{eq:softtriple}
\ell(\mathbf{x}_{i})=-\log \frac{\exp \left(\lambda\left(\mathcal{S}_{i, y_{i}}^{\prime}-\delta\right)\right)}{\exp \left(\lambda\left(\mathcal{S}_{i, y_{i}}^{\prime}-\delta\right)\right)+\sum_{j} \exp \left(\lambda \mathcal{S}_{i, j}^{\prime}\right)}
\end{equation}

 \cite{qian2019softtriple} used a regularizer to set the center size in an adaptive way to avoid overfitting. The regularizer requires a matrix that grows quadratically in the number of classes. This made it intractable for us to use it in our task as we have 10k classes. Therefore, we chose to omit the regularizer and go for a large number of centers. In addition, we set $\gamma=1$ in Eq. \ref{eq:similarity}, since our experiments showed that when $\lambda$ is large, it dominates the results and the effect of $\gamma$ can be ignored. 

% Our ablation study reveals that models without $\gamma$ show comparable performance as models that have it.

\section{Experiments}
\label{sec:experiments}

We ran the Montreal Forced Aligner \cite{mcauliffe2017montreal} on the Librispeech \cite{panayotov2015librispeech} dataset to extract the alignments between the audio and the annotated text. We used Librispeech training sets (train-clean-100, train-clean-360, train-other-500) to train the model and the dev sets (dev-clean, dev-other) to validate the training process. The models were then trained to output the corresponding top 10k annotated words for a given audio. Added to that, we applied babble noise at uniform random SNRs in the [5, 15] dB range. 

% After training the model, we test its performance on internal data containing English utterances. This causes a mismatch in the distribution of the training data and the test utterances. To combat this issue, we use a small subset of the internal English dataset for validation in addition to Librispeech's dev sets. Specifically, after training the model on Librispeech for one epoch, we run the model on the Librispeech dev sets and check if the loss on the dev set is lower than the current minimum value. If the new loss is less than the current minimum loss, the model is run on the internal validation set. This process is repeated until the performance on the Librispeech dev sets no longer improves. The model with the best performance on the internal validation set is then saved for testing. Using this training procedure, we obtain a model that generalizes well on Librispeech and performs well on the QbyE task.

% \subsection{Experimental Setup}
% \label{setup}

Our system was evaluated on both an internal dataset, and the publicly available Hey-Snips dataset \cite{coucke2019efficient}. The internal datasets contained utterances of 8 English keywords and 11 Korean keywords from 50 speakers each shown in the table. We used a 24-hour long recording of TV programs as negative samples. Keywords also serve as negatives if they are different from the enrolled keywords. The internal datasets were concatenated together to mimic the on-device situation where the signals are input as a stream. 

\begin{table}[tbh]
\label{tab:keywords}
\resizebox{\columnwidth}{!}{%
\begin{tabular}{|l|l|l|l|l|}
\hline
\multirow{2}{*}{\begin{tabular}[c]{@{}l@{}}English \\ 8 Total\end{tabular}} & LG Styler & LG Washer & LG Dryer & Hey LG \\ \cline{2-5} 
 & LG Fridge & LG Puricare & LG Oven  & Hey Cloi \\ \hline
\multirow{2}{*}{\begin{tabular}[c]{@{}l@{}}Korean\\ 11 Total\end{tabular}} & \multicolumn{4}{c|}{Same 8 English keywords in Korean} \\ \cline{2-5} 
 & Yeongmi Ya & LG Air Purifier & \multicolumn{2}{l|}{Yeongmi Yeongmi} \\ \hline
\end{tabular}
}
\end{table}

The Hey-Snips dataset contains examples of one keyword "Hey Snips" along with examples of general sentences. We selected all the speakers with 10 keyword utterances (the maximum number recorded for an individual in that dataset) as positives, and all the general sentences as negatives, this yielded 98, 44, and 40 speakers with positive keywords, and 44,860, 20,181, and 20,543 sentences as negative samples in the train, dev, and test sets respectively.

In the test phase, 3 utterances were randomly picked as enrollments for each speaker. We add 10 dB babble noise\footnote{Babble noise samples were obtained from github.com/microsoft/MS-SNSD} to all the test utterances. The babble noise used during training is different from the one used during testing. For a given query, the cosine distance is used to compare the similarity between the query embedding and the 3 enrolled embeddings. The minimum distance is selected and compared with a threshold value to make the detection decision. For the internal datasets, if the system outputs a detection, a 2 second suppression filter is applied to prevent the system from outputting another one. For Hey-Snips, the FRR is calculated at the utterance level. 

% \textcolor{blue}{In order to compare the proposed CNN approaches to a baseline
% DNN KWS system, we selected fourteen phrases2
% and collected
% about 10K–15K utterances containing each of these phrases.
% We also collected a much larger set of approximately 396K utterances which do not contain any of the keywords and are thus
% used as ‘negative’ training data. The utterances were then randomly split into training, development, and evaluation sets in
% the ratio of 80:5:15, respectively.
% Next, we created noisy training and evaluation sets by artificially adding car and cafeteria noise at SNRs randomly sampled between [-5dB, +10dB] to the clean data sets. Models
% are trained in noisy conditions, and evaluated in both clean and
% noisy conditions.
% KWS performance is measured by plotting a receiver operating curve (ROC), which calculates the false reject (FR) rate
% per false alarm (FA) rate. The lower the FR per FA rate is the
% better. The KWS system threshold is selected to correspond to
% 1 FA per hour of speech on this set.}

% \subsection{Model Selection with Hyper-Parameter Search}
% We use Adam optimizer to conduct the gradient search for our model. In the mean time, we use Asynchronous Successive Halving Algorithm (ASHA) \cite{li2018massively} as scheduler for a massive (more than 1000 experiments) parallel hyper-parameter search for hyper-parameters such as number of layers, hidden units, CNN filter stride, window size, etc. The network architecture presented in Section \ref{sec:system} is among the top performed ones selected from the ASHA scheduler.

\section{Results}
\label{sec:results}
% Some initial results are as follows, we will conduct more experiments to finalize this section later. 

Based on our network structure described in Section \ref{sec:system}, we designed two models to cater to various home appliances. One with an encoder containing 292k parameters and the other one with 582k parameters. In both models, we use 20 heads in MH-E and 15 heads in NMH-A. For the last fully connected layer, the number of centers is set to 6 with $\lambda=70$ and $\delta=0.04$. The base encoder for the 292k model is a 4 layer GRU with hidden size 100 while the 582k model is a 6 layer GRU with hidden size 120. These hyperparameters were selected from a massive (more than 1000 experiments) parallel hyperparameter search using the Asynchronous Successive Halving Algorithm (ASHA).

\subsection{Evaluation Results}
\label{internal_data}
We conducted our experiments on clean data and the data augmented with 10 dB babble noise for both Korean and English datasets. There is also a significant gap in FRR between clean data and data augmented with babble noise. It is worth noting that even though our models were trained on English utterances, the models perform better on clean Korean data than clean English data. This can be attributed to the average utterance length which is around 3 seconds in Korean compared to 1.5 seconds in English. We empirically observed that longer utterances were easier to detect than shorter ones. The performance on babble data is roughly similar across both languages. This shows the versatile nature of our models and their ability to adapt to various data distributions.

\begin{figure}[htb]
  \centering
  \includegraphics[width=8.5cm]{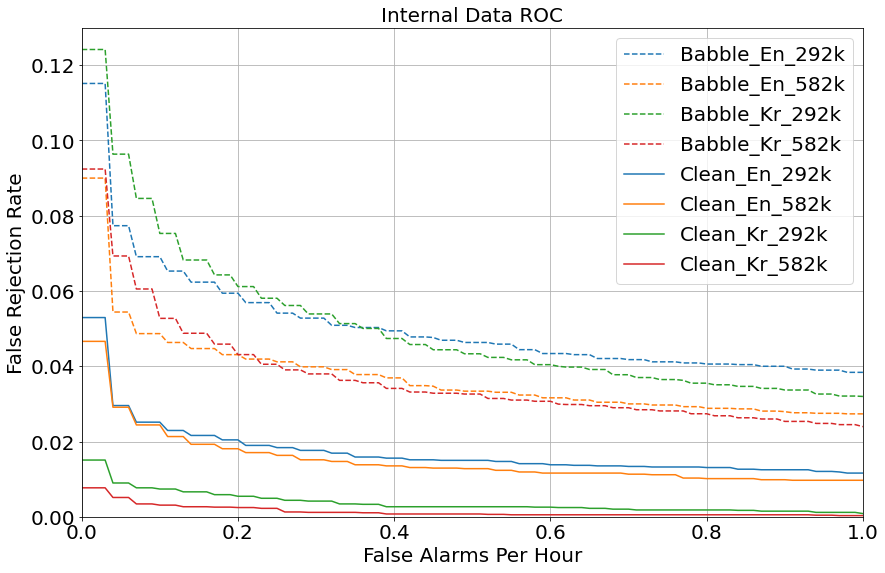}
%  \vspace{2.0cm}
\caption{ROC on Internal English and Korean Data}
\label{fig:internal ROC}
\end{figure}

\begin{figure}[htb]
  \centering
  \includegraphics[width=8.5cm]{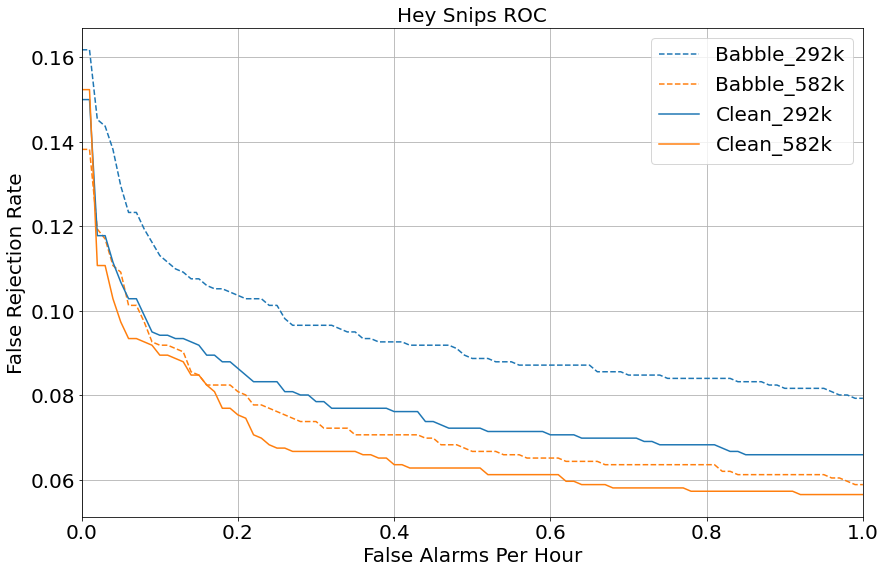}
%  \vspace{2.0cm}
\caption{Hey-Snips ROC with clean data versus babble noise. This plot is created by using all 102.88 hours of Hey-Snips negatives and tested on 182 speakers.}
\label{fig:hey snips}
\end{figure}

% \subsection{Hey-Snips}
% \label{hey_snips}
 We also ran experiments on Hey-Snips with clean data and data with babble noise. The results in Fig. \ref{fig:internal ROC} and Fig. \ref{fig:hey snips} show that the 582k model outperforms the 292k model across datasets and noise which showcases the large model's discriminative ability. In addition, we observe that the performance of our models on Hey-Snips is not as good as the internal dataset. This might be due to the fact that Hey-Snips has relatively shorter utterances. 
 
%  The results are shown in Fig. \ref{fig:hey snips}. 

% For lower FAR, the 600k model outperforms the 300k model.. etc....

\subsection{Baseline Comparison and Ablation Study}
\label{ablation}
We used the method proposed from \cite{chen2015query} as our baseline system. Specifically, we used the best-configured GRU layers with the last 5 time steps from the embedding vector to compare the distance. The False Rejection Rate (FRR) at 0.3 FA/hour on both the internal English (En) as well as the Hey-Snips (HS) dataset results for models with parameters around 300k (Small) and 600k (Large) is shown in Table \ref{tab:comparison}. The results illustrate that our model outperforms the baseline system by a large margin.

\begin{table}[]
\center
\resizebox{\columnwidth}{!}{
\begin{tabular}{lcccc}
         & \textbf{En (Clean)} & \textbf{En (Babble)} & \textbf{HS (Clean)} & \textbf{HS (Babble)} \\
      \hline
Baseline (Small) & 17.87           & 11.43            & 12.09             & 11.70              \\
Baseline (Large) & 13.29           & 8.76             & 11.70             & 9.58               \\
Ours (Small)     & 1.76            & 5.28             & 7.85              & 9.65               \\
\textbf{Ours (Large)}   & \textbf{1.51}            & \textbf{3.99}             & \textbf{6.67}              & \textbf{7.38} \\
\hline
\end{tabular}
}
\caption{Baseline Comparison}
\label{tab:comparison}
\end{table}

% \begin{table}[]
% \center
% \resizebox{0.7\columnwidth}{!}{
% \begin{tabular}{lcc|cc}
%                 & \multicolumn{2}{c|}{English} & \multicolumn{2}{c}{Hey-Snips} \\
%                 & Clean        & Babble        & Clean         & Babble        \\ \hline
% Baseline (296k) & 17.87        & 11.43         & 12.09         & 11.70         \\
% Ours (296k)     & 1.76         & 5.28          & 7.85          & 9.65          \\
% Baseline (296k) & 13.29        & 8.76          & 11.70         & 9.58          \\
% Ours (586k)     & 1.51         & 3.99          & 6.67          & 7.38         
% \end{tabular}
% }
% \label{tab:comparison}
% \caption{False Rejection Rate (FRR) at 0.3 FA/hour on both the internal English (En) as well as the Hey-Snips (HS) dataset. The results above indicate that our method outperforms the baseline system \cite{chen2014small} on both datasets.}
% \end{table}

Additionally, to demonstrate the effectiveness of every module in our architecture, we conducted an ablation study by removing the individual components and measuring their performance on the internal English dataset with 10 dB babble noise. The results are shown in Table \ref{tab:ablation}. 

(1) and (6) show the effectiveness of softtriple as substituting it for a softmax loss leads to more than 8.5\% increase in FRR for the small model and 5\% increase for the large model. (2) and (6) show that adding a Multi-Head attention feature Extractor (MH-E) decreases FRR by around 2\%. (3) and (4) compare  an attention mechanism that uses tanh activation \cite{rahman2018attention} to our non-normalized version of Multi-Head Aggregator (MH-A). \cite{rahman2018attention} adds more parameters while achieving comparable performance with the non-normalized MH-A. Comparing (6) to (3) and (4) shows that Normalizing the Multi-Head Aggregator weights ((NMH-A) results in more than 2\% reduction in FRR. Finally, (5) and (6) show that increasing the number of heads in the multi-head aggregator (1 head vs 15 heads) further boosts the performance which indicates that the model benefits from the multidimensional aspect of the attention mechanism.

\begin{table}[]
\resizebox{\columnwidth}{!}{
\begin{tabular}{llcc}
& \textbf{Model}        & \textbf{FRR (Small)} & \textbf{FRR (Large)} \\ \hline
(1) & MH-E + NMH-A + Softmax         & 13.91                 & 9.03                  \\
(2) &  NMH-A + Softtriple                & 7.53                  & 6.04                  \\
(3) &  MH-E + Tanh\cite{rahman2018attention} + Softtriple        & 7.22                  & 7.99                  \\
(4) & MH-E + MH-A + Softtriple          & 7.51                  & 6.10                  \\
(5) & MH-E + NMH-A (1 head) + Softtriple         & 7.03                  & 4.91                  \\
(6) &  \textbf{MH-E + NMH-A + Softtriple} & \textbf{5.28}         & \textbf{3.99}         \\ \hline

\end{tabular}
}
\caption{Ablation study for various components in our model. MH-E is the multi-head attention feature extractor, NMH-A is the multi-head attention feature extractor with $L_2$ normalization, and MH-A is the multi-head attention feature aggregator without $L_2$ normalization. FRR is reported at 0.3 FA/hour \label{tab:ablation}}
\end{table}

% The model obtains 15\% higher recall on clean data than babble noise. This is due to the challenging nature of babble noise where the model has to separate the background conversations from the test utterance. To understand which keywords were more challenging for the model, the ROC for all the keywords in the English internal dataset with babble noise is shown in figure \ref{fig:keyword ROC}. 

% The model has the best recall on ``Hey Cloi" and ``Hey LG" probably because the beginning of the utterance is different from other keywords since they start with ``LG". On the other hand, the model obtains the worst recall on ``LG Styler" and ``LG Puricare".  

\section{Conclusion}
\label{sec:conclusion}

We propose a neural architecture to tackle the QbyE user-defined KWS task which makes use of a multi-headed attention feature extractor and a normalized multi-head attention feature aggregator mechanism. We also demonstrate the effectiveness of softtriple loss which was originally tested on a small scale image classification task and compare the performance of our system to a baseline. Finally, we conduct an ablation study to highlight the contribution of every component in our architecture and establish a baseline on the Hey-Snips dataset for future comparison.

\vfill\pagebreak

\bibliographystyle{IEEEbib}
\bibliography{strings,refs}

\end{document}